%% file: wsd.tex
\documentclass[11pt,a4paper]{article}
\usepackage[hyperref]{emnlp-ijcnlp-2019}
\usepackage{times}
\usepackage{latexsym}

\usepackage{url}

\usepackage{amsfonts}
\usepackage{amsmath}
\usepackage{graphicx}
\usepackage{subcaption}

\aclfinalcopy 

\title{Improved Word Sense Disambiguation \\ Using Pre-Trained Contextualized Word Representations}

\author{Christian Hadiwinoto, Hwee Tou Ng, \and Wee Chung Gan \\
  Department of Computer Science, National University of Singapore \\
  {\tt christian.hadiwinoto@u.nus.edu, nght@comp.nus.edu.sg,} \\
  {\tt gan\_weechung@u.nus.edu} \\}

\date{}

\begin{document}
\maketitle
\begin{abstract}
Contextualized word representations are able to give different representations for the same word in different contexts, and they have been shown to be effective in downstream natural language processing tasks, such as question answering, named entity recognition, and sentiment analysis. However, evaluation on word sense disambiguation (WSD) in prior work shows that using contextualized word representations does not outperform the state-of-the-art approach that makes use of non-contextualized word embeddings. In this paper, we explore different strategies of integrating pre-trained contextualized word representations and our best strategy achieves accuracies exceeding the best prior published accuracies by significant margins on multiple benchmark WSD datasets. We make the source code available at \url{https://github.com/nusnlp/contextemb-wsd}.
\end{abstract}

\section{Introduction}
\label{sec:intro}

Word sense disambiguation (WSD) automatically assigns a pre-defined sense to a word in a text. Different senses of a word reflect different meanings a word has in different contexts. Identifying the correct word sense given a context is crucial in natural language processing (NLP). Unfortunately, while it is easy for a human to infer the correct sense of a word given a context, it is a challenge for NLP systems. As such, WSD is an important task and it has been shown that WSD helps downstream NLP tasks, such as machine translation \cite{chan_word_2007} and information retrieval \cite{zhong_word_2012}.

A WSD system assigns a sense to a word by taking into account its context, comprising the other words in the sentence. This can be done through discrete word features, which typically involve surrounding words and collocations trained using a classifier \cite{lee_supervised_2004,ando_applying_2006,chan_nus-pt:_2007,zhong_it_2010}. The classifier can also make use of continuous word representations of the surrounding words \cite{taghipour_semi-supervised_2015,iacobacci_embeddings_2016}. Neural WSD systems \cite{kageback_word_2016,raganato_neural_2017} feed the continuous word representations into a neural network that captures the whole sentence and the word representation in the sentence. However, in both approaches, the word representations are independent of the context.

Recently, pre-trained contextualized word representations \cite{melamud_context2vec:_2016,mccann_learned_2017,peters_deep_2018,devlin_bert:_2019} have been shown to improve downstream NLP tasks. Pre-trained contextualized word representations are obtained through neural sentence encoders trained on a huge amount of raw texts. When the resulting sentence encoder is fine-tuned on the downstream task, such as question answering, named entity recognition, and sentiment analysis, with much smaller annotated training data, it has been shown that the trained model, with the pre-trained sentence encoder component, achieves new state-of-the-art results on those tasks.

While demonstrating superior performance in downstream NLP tasks, pre-trained contextualized word representations are still reported to give lower accuracy compared to approaches that use non-contextualized word representations \cite{melamud_context2vec:_2016,peters_deep_2018} when evaluated on WSD. This seems counter-intuitive, as a neural sentence encoder better captures the surrounding context that serves as an important cue to disambiguate words. In this paper, we explore different strategies of integrating pre-trained contextualized word representations for WSD. Our best strategy outperforms prior methods of incorporating pre-trained contextualized word representations and achieves new state-of-the-art accuracy on multiple benchmark WSD datasets.

The following sections are organized as follows. Section \ref{sec:related} presents related work. Section \ref{sec:pretrain} describes our pre-trained contextualized word representation. Section \ref{sec:incorporating} proposes different strategies to incorporate the contextualized word representation for WSD. Section \ref{sec:exp} describes our experimental setup. Section \ref{sec:results} presents the experimental results. Section \ref{sec:discussion} discusses the findings from the experiments. Finally, Section \ref{sec:conclusion} presents the conclusion.

\section{Related Work}
\label{sec:related}

Continuous word representations in real-valued vectors, or commonly known as word embeddings, have been shown to help improve NLP performance. Initially, exploiting continuous representations was achieved by adding real-valued vectors as classification features \cite{turian_word_2010}. \citet{taghipour_semi-supervised_2015} fine-tuned non-contextualized word embeddings by a feed-forward neural network such that those word embeddings were more suited for WSD. The fine-tuned embeddings were incorporated into an SVM classifier. \citet{iacobacci_embeddings_2016} explored different strategies of incorporating word embeddings and found that their best strategy involved exponential decay that decreased the contribution of surrounding word features as their distances to the target word increased.

The neural sequence tagging approach has also been explored for WSD. \citet{kageback_word_2016} proposed bidirectional long short-term memory (LSTM) \cite{hochreiter_long_1997} for WSD. They concatenated the hidden states of the forward and backward LSTMs and fed the concatenation into an affine transformation followed by softmax normalization, similar to the approach to incorporate a bidirectional LSTM adopted in sequence labeling tasks such as part-of-speech tagging and named entity recognition \cite{ma_end--end_2016}. \citet{raganato_neural_2017} proposed a self-attention layer on top of the concatenated bidirectional LSTM hidden states for WSD and introduced multi-task learning with part-of-speech tagging and semantic labeling as auxiliary tasks. However, on average across the test sets, their approach did not outperform SVM with word embedding features. Subsequently, \citet{luo_incorporating_2018} proposed the incorporation of glosses from WordNet in a bidirectional LSTM for WSD, and reported better results than both SVM and prior bidirectional LSTM models.

A neural language model (LM) is aimed at predicting a word given its surrounding context. As such, the resulting hidden representation vector captures the context of a word in a sentence. \citet{melamud_context2vec:_2016} designed \textit{context2vec}, which is a one-layer bidirectional LSTM trained to maximize the similarity between the hidden state representation of the LSTM and the target word embedding. \citet{peters_deep_2018} designed ELMo, which is a two-layer bidirectional LSTM language model trained to predict the next word in the forward LSTM and the previous word in the backward LSTM. In both models, WSD was evaluated by nearest neighbor matching between the test and training instance representations. However, despite training on a huge amount of raw texts, the resulting accuracies were still lower than those achieved by WSD approaches with pre-trained non-contextualized word representations.

End-to-end neural machine translation (NMT) \cite{sutskever_sequence_2014,bahdanau_neural_2015} learns to generate an output sequence given an input sequence, using an encoder-decoder model. The encoder captures the contextualized representation of the words in the input sentence for the decoder to generate the output sentence. Following this intuition, \citet{mccann_learned_2017} trained an encoder-decoder model on parallel texts and obtained pre-trained contextualized word representations from the encoder.

\section{Pre-Trained Contextualized Word Representation}
\label{sec:pretrain}

The contextualized word representation that we use is BERT \cite{devlin_bert:_2019}, which is a bidirectional transformer encoder model \cite{vaswani_attention_2017} pre-trained on billions of words of texts. There are two tasks on which the model is trained, i.e., masked word and next sentence prediction. In both tasks, prediction accuracy is determined by the ability of the model to understand the context.

A transformer encoder computes the representation of each word through an attention mechanism with respect to the surrounding words. Given a sentence $x^n_1$ of length $n$, the transformer computes the representation of each word $x_i$ through a multi-head attention mechanism, where the query vector is from $x_i$ and the key-value vector pairs are from the surrounding words $x_{i'}$ ($1 \le i' \le n$). The word representation produced by the transformer captures the contextual information of a word.

The attention mechanism can be viewed as mapping a query vector $\mathbf{q}$ and a set of key-value vector pairs $(\mathbf{k}, \mathbf{v})$ to an output vector. The attention function $A(\cdot)$ computes the output vector which is the weighted sum of the value vectors and is defined as:
\begin{align}
\label{eq:att}
A(\mathbf{q}, \mathbf{K}, \mathbf{V}, \rho) &= \sum_{(\mathbf{k}, \mathbf{v}) \in (\mathbf{K}, \mathbf{V})}{\alpha(\mathbf{q}, \mathbf{k}, \rho) \mathbf{v}} \\
\alpha(\mathbf{q}, \mathbf{k}, \rho) &= \frac{\exp(\rho \mathbf{k}^\top \mathbf{q})}{\sum_{\mathbf{k}' \in \mathbf{K}}{\exp(\rho \mathbf{k}'^\top \mathbf{q})}}
\end{align}
where $\mathbf{K}$ and $\mathbf{V}$ are matrices, containing the key vectors and the value vectors of the words in the sentence respectively, and $\alpha(\mathbf{q}, \mathbf{k}, \rho)$ is a scalar attention weight between $\mathbf{q}$ and $\mathbf{k}$, re-scaled by a scalar $\rho$.

Two building blocks for the transformer encoder are the multi-head attention mechanism and the position-wise feed-forward neural network (FFNN). The multi-head attention mechanism with $H$ heads leverages the attention function in Equation \ref{eq:att} as follows:
\begin{align}
\label{eq:nmt_transf_mhead}
\text{MH}(\mathbf{q}, \mathbf{K}, \mathbf{V}, \rho) &= \mathbf{W}_{\text{MH}} \bigoplus_{\eta=1}^{H} \text{head}_\eta(\mathbf{q}, \mathbf{K}, \mathbf{V}, \rho) \\
\label{eq:nmt_transf_head}
\text{head}_\eta(\mathbf{q}, \mathbf{K}, \mathbf{V}, \rho) &= A(\mathbf{W}^\mathbf{Q}_\eta \mathbf{q}, \mathbf{W}^\mathbf{K}_\eta \mathbf{K}, \mathbf{W}^\mathbf{V}_\eta \mathbf{V}, \rho)
\end{align}
where $\oplus$ denotes concatenation of vectors, $\mathbf{W}_\text{MH} \in \mathbb{R}^{d_\text{model} \times Hd_\mathbf{v}}$, $\mathbf{W}^\mathbf{Q}_\eta, \mathbf{W}^\mathbf{K}_\eta \in \mathbb{R}^{d_\mathbf{k} \times d_\text{model}}$, and $ \mathbf{W}^\mathbf{V}_\eta \in \mathbb{R}^{d_\mathbf{v} \times d_\text{model}}$. The input vector $\mathbf{q} \in \mathbb{R}^{d_\text{model}}$ is the hidden vector for the ambiguous word, while input matrices $\mathbf{K}, \mathbf{V} \in \mathbb{R}^{d_\text{model} \times n}$ are the concatenation of the hidden vectors of all words in the sentence. For each attention head, the dimension of both the query and key vectors is $d_\mathbf{k}$ while the dimension of the value vector is $d_\mathbf{v}$. The encoder model dimension is $d_\text{model}$.

The position-wise FFNN performs a non-linear transformation on the attention output corresponding to each input word position as follows:
\begin{equation}
\label{eq:nmt_transf_ffnn}
\text{FF}(\mathbf{u}) = \mathbf{W}_\text{FF2} (\max(0, \mathbf{W}_\text{FF1} \mathbf{u} + \mathbf{b}_\text{FF1})) + \mathbf{b}_\text{FF2}
\end{equation}
in which the input vector $\mathbf{u} \in \mathbb{R}^{d_\text{model}}$ is transformed to the output vector with dimension $d_\text{model}$ via a series of linear projections with the ReLU activation function.

For the hidden layer $l$ ($1 \le l \le L$), the self-attention sub-layer output $\mathbf{f}^l_i$ is computed as follows:
\begin{align*}
\mathbf{f}^l_i &= \text{LayerNorm}(\mathbf{\chi}^l_{\mathbf{h},i} + \mathbf{h}^{l-1}_i) \\
\mathbf{\chi}^l_{\mathbf{h},i} &= \text{MH}^l_\mathbf{h}(\mathbf{h}^{l-1}_{i}, \mathbf{h}^{l-1}_{1:n}, \mathbf{h}^{l-1}_{1:n}, \dfrac{1}{\sqrt{d_\text{v}}})
\end{align*}
where LayerNorm refers to layer normalization \cite{DBLP:journals/corr/BaKH16} and the superscript $l$ and subscript $\mathbf{h}$ indicate that each encoder layer $l$ has an independent set of multi-head attention weight parameters (see Equations \ref{eq:nmt_transf_mhead} and \ref{eq:nmt_transf_head}). The input for the first layer is $\mathbf{h}^0_i = \mathbf{E}(x_i)$, which is the non-contextualized word embedding of $x_i$.

The second sub-layer consists of the position-wise fully connected FFNN, computed as:
\begin{equation*}
\mathbf{h}^l_i = \text{LayerNorm}(\text{FF}^{l}_\mathbf{h}(\mathbf{f}^{l}_{i}) + \mathbf{f}^{l}_{i})
\end{equation*}
where, similar to self-attention, an independent set of weight parameters in Equation \ref{eq:nmt_transf_ffnn} is defined in each layer.

\section{Incorporating Pre-Trained Contextualized Word Representation}
\label{sec:incorporating}

As BERT is trained on the masked word prediction task, which is to predict a word given the surrounding (left and right) context, the pre-trained model already captures the context. In this section, we describe different techniques of leveraging BERT for WSD, broadly categorized into nearest neighbor matching and linear projection of hidden layers.

\subsection{Nearest Neighbor Matching}
\label{sec:incorporating_nearest}

A straightforward way to disambiguate word sense is through 1-nearest neighbor matching. We compute the contextualized representation of each word in the training data and the test data through BERT. Given a hidden representation $\mathbf{h}^L_{i}$ at the $L$-th layer for word $x_i$ in the test data, nearest neighbor matching finds a vector $\mathbf{h^*}$ in the $L$-th layer from the training data such that
\begin{equation}
\label{eq:nearest}
    \mathbf{h^*} = \arg\max_{\mathbf{h'}}{\cos(\mathbf{h}^L_{i},\mathbf{h'})}
\end{equation}
where the sense assigned to $x_i$ is the sense of the word whose contextualized representation is $\mathbf{h^*}$. This WSD technique is adopted in earlier work on contextualized word representations \cite{melamud_context2vec:_2016,peters_deep_2018}.

\subsection{Linear Projection of Hidden Layers}
\label{sec:incorporating_linear}

\begin{figure*}
\centering{
\begin{subfigure}[b]{0.4\textwidth}
\centering
\includegraphics[scale=0.8]{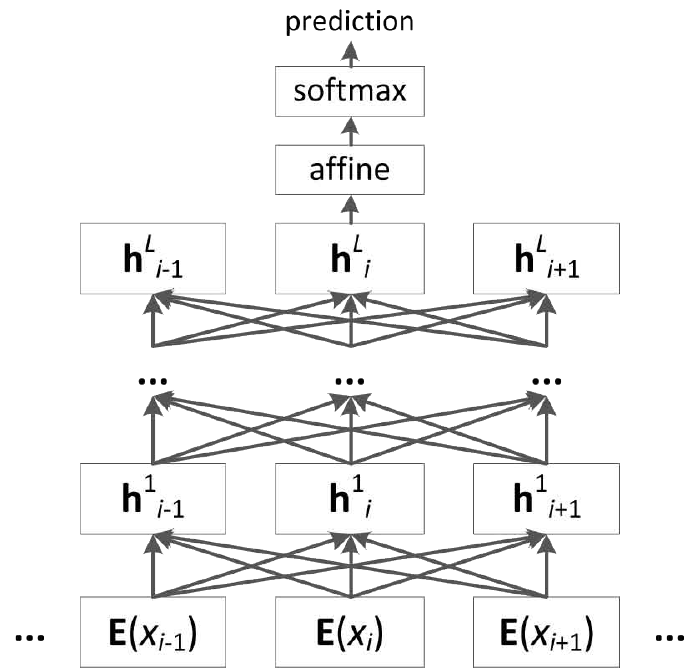}
\caption{~}
\end{subfigure}
\begin{subfigure}[b]{0.4\textwidth}
\centering
\includegraphics[scale=0.8]{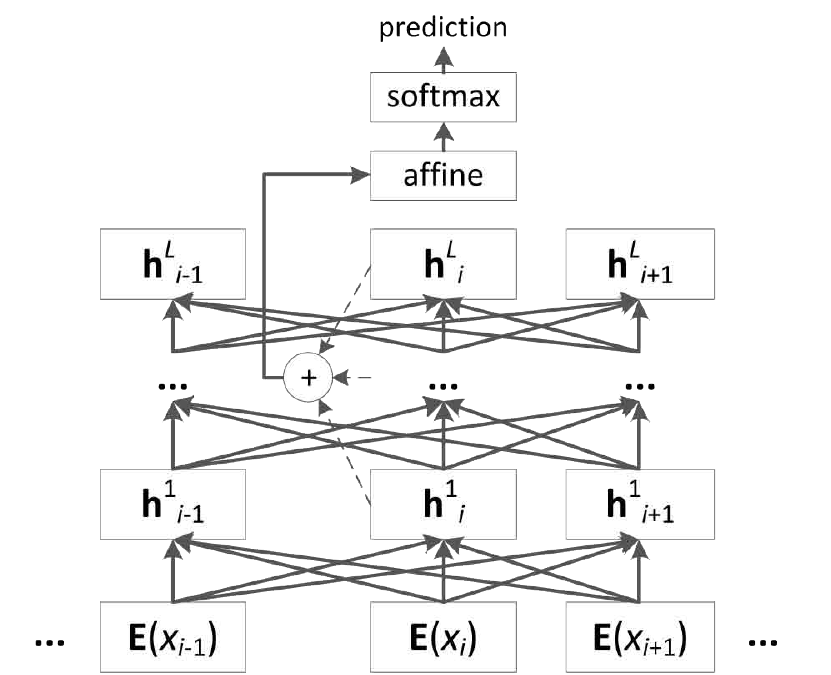}
\caption{~}
\end{subfigure}
}
\caption{\label{fig:linear} Illustration of WSD models by linear projection of (a) the last layer and (b) the weighted sum of all layers.}
\end{figure*}

Apart from nearest neighbor matching, we can perform a linear projection of the hidden vector $\mathbf{h}_i$ by an affine transformation into an output sense vector, with its dimension equal to the number of senses for word $x_i$. The output of this affine transformation is normalized by softmax such that all its values sum to $1$. Therefore, the predicted sense $\mathbf{s}_i$ of word $x_i$ is formulated as
\begin{equation}
\label{eq:linear}
\mathbf{s}_i = \text{softmax}(\mathbf{W}^{\text{lexelt}(x_i)}\mathbf{h}_i + \mathbf{b}^{\text{lexelt}(x_i)})
\end{equation}
where $\mathbf{s}_i$ is a vector of predicted sense distribution for word $x_i$, while $\mathbf{W}^{\text{lexelt}(x_i)}$ and $\mathbf{b}^{\text{lexelt}(x_i)}$ are respectively the projection matrix and bias vector specific to the lexical element (lexelt) of word $x_i$, which consists of its lemma and optionally its part-of-speech tag. We choose the sense corresponding to the element of $\mathbf{s}_i$ with the maximum value.

Training the linear projection model is done by the back-propagation algorithm, which updates the model parameters to minimize a cost function. Our cost function is the negative log-likelihood of the softmax output value that corresponds to the tagged sense in the training data. In addition, we propose two {\it novel} ways of incorporating BERT's hidden representation vectors to compute the sense output vector, which are described in the following sub-subsections.

\subsubsection{Last Layer Projection}
\label{sec:incorporating_linear_last}

Similar to the nearest neighbor matching model, we can feed the hidden vector of BERT in the last layer, $\mathbf{h}^L_i$, into an affine transformation followed by softmax. That is, $\mathbf{h}_i$ in Equation \ref{eq:linear} is instantiated by $\mathbf{h}^L_i$. The last layer projection model is illustrated in Figure \ref{fig:linear}(a).

\subsubsection{Weighted Sum of Hidden Layers}
\label{sec:incorporating_linear_lw}

BERT consists of multiple layers stacked one after another. Each layer carries a different representation of word context. Taking into account different hidden layers may help the WSD system to learn from different context information encoded in different layers of BERT.

To take into account all layers, we compute the weighted sum of all hidden layers, $\mathbf{h}^l_i$, where $1 \le l \le L$, corresponding to a word position $i$, through attention mechanism. That is, $\mathbf{h}_i$ in Equation \ref{eq:linear} is replaced by the following equation:
\begin{equation}
    \mathbf{h}_i = A(\mathbf{m}, \mathbf{W}^\mathbf{s} \mathbf{h}^{1:L}_i, \mathbf{h}^{1:L}_i, 1)
\end{equation}
where $\mathbf{m} \in \mathbb{R}^{d_\text{model}}$ is a projection vector to obtain scalar values with the key vectors. The model with weighted sum of all hidden layers is illustrated in Figure \ref{fig:linear}(b).

\subsubsection{Gated Linear Unit}
\label{sec:incorporating_linear_glu}

While the contextualized representations in the BERT hidden layer vectors are features that determine the word sense, some features are more useful than the others. As such, we propose filtering the vector values by a gating vector whose values range from $0$ to $1$. This mechanism is known as the gated linear unit (GLU) \cite{dauphin_language_2017}, which is formulated as
\begin{equation}
    \text{GLU}(\mathbf{h}) = (\mathbf{h} + \mathbf{W}^\mathbf{h}\mathbf{h} + \mathbf{b}^\mathbf{h}) \odot \sigma(\mathbf{W}^\mathbf{g}\mathbf{h} + \mathbf{b}^\mathbf{g})
\end{equation}
where $\mathbf{W}^\mathbf{h}$ and $\mathbf{W}^\mathbf{g}$ are separate projection matrices and $\mathbf{b}^\mathbf{h}$ and $\mathbf{b}^\mathbf{g}$ are separate bias vectors. The symbols $\sigma(\cdot)$ and $\odot$ denote the sigmoid function and element-wise vector multiplication operation respectively.

GLU transforms the input vector $\mathbf{h}$ by feeding it to two separate affine transformations. The second transformation is used as the sigmoid gate to filter the input vector, which is summed with the vector after the first affine transformation.

\section{Experimental Setup}
\label{sec:exp}

We conduct experiments on various WSD tasks. The description and the statistics for each task are given in Table \ref{tab:data}. For English, a lexical element (lexelt) is defined as a combination of lemma and part-of-speech tag, while for Chinese, it is simply the lemma, following the OntoNotes setup.

We exploit English BERT$_\text{BASE}$ for the English tasks and Chinese BERT for the Chinese task. We conduct experiments with different strategies of incorporating BERT as described in Section \ref{sec:incorporating}, namely 1-nearest neighbor matching (\textbf{1-nn}) and linear projection. In the latter technique, we explore strategies including \textbf{simple} last layer projection, layer weighting (\textbf{LW}), and gated linear unit (\textbf{GLU}).

In the linear projection model, we train the model by the Adam algorithm \cite{kingma_adam:_2015} with a learning rate of $10^{-3}$. The model parameters are updated per mini-batch of 16 sentences. As update progresses, we pick the best model parameters from a series of neural network updates based on accuracy on a held-out development set, disjoint from the training set.

The state-of-the-art supervised WSD approach takes into account features from the neighboring sentences, typically one sentence to the left and one to the right apart from the current sentence that contains the ambiguous words. We also exploit this in our model, as BERT supports inputs with multiple sentences separated by a special \texttt{[SEP]} symbol.

\input{tables/traindata.tex}

For English all-words WSD, we train our WSD model on SemCor \cite{miller_using_1994}, and test it on Senseval-2 (SE2), Senseval-3 (SE3), SemEval 2013 task 12 (SE13), and SemEval 2015 task 13 (SE15). This common benchmark, which has been annotated with WordNet-3.0 senses \cite{raganato_word_2017}, has recently been adopted in English all-words WSD. Following \cite{raganato_neural_2017}, we choose SemEval 2007 Task 17 (SE07) as our development data to pick the best model parameters after a number of neural network updates, for models that require back-propagation training.

We also evaluate on Senseval-2 and Senseval-3 English lexical sample tasks, which come with pre-defined training and test data. For each word type, we pick 20\% of the training instances to be our development set and keep the remaining 80\% as the actual training data. Through this development set, we determine the number of epochs to use in training. We then re-train the model with the whole training dataset using the number of epochs identified in the initial training step.

While WSD is predominantly evaluated on English, we are also interested in evaluating our approach on Chinese, to evaluate the effectiveness of our approach in a different language. We use OntoNotes Release 5.0\footnote{LDC2013T19}, which contains a number of annotations including word senses for Chinese. We follow the data setup of \citet{pradhan_towards_2013} and conduct an evaluation on four genres, i.e., broadcast conversation (BC), broadcast news (BN), magazine (MZ), and newswire (NW), as well as the concatenation of all genres. While the training and development datasets are divided into genres, we train on the concatenation of all genres and test on each individual genre.

\input{tables/results_aw.tex}

For Chinese WSD evaluation, we train IMS \cite{zhong_it_2010} on the Chinese OntoNotes dataset as our baseline. We also incorporate pre-trained non-contextualized Chinese word embeddings as IMS features \cite{taghipour_semi-supervised_2015,iacobacci_embeddings_2016}. The pre-trained word embeddings are obtained by training the \textit{word2vec} skip-gram model on Chinese Gigaword Fifth Edition\footnote{LDC2011T13}, which after automatic word segmentation contains approximately 2 billion words. Following \cite{taghipour_semi-supervised_2015}, we incorporate the embedding features of words within a window surrounding the target ambiguous word. In our experiments, we take into account 5 words to the left and right.

\section{Results}
\label{sec:results}

We present our experimental results and compare them with prior baselines.

\subsection{English All-Words Tasks}
\label{sec:results_aw}

For English all-words WSD, we compare our approach with three categories of prior approaches. Firstly, we compare our approach with the supervised SVM classifier approach, namely IMS \cite{zhong_it_2010}. We compare our approach with both the original IMS without word embedding features and IMS with non-contextualized word embedding features, that is, \textit{word2vec} with exponential decay \cite{iacobacci_embeddings_2016}. We also compare with SupWSD \cite{papandrea_supwsd:_2017}, which is an alternative implementation of IMS. Secondly, we compare our approach with the neural WSD approaches that leverage bidirectional LSTM (bi-LSTM). These include the bi-LSTM model with attention trained jointly with lexical semantic labeling task \cite{raganato_neural_2017} (BiLSTMatt+LEX) and the bi-LSTM model enhanced with gloss representation from WordNet (GAS). Thirdly, we show comparison with prior contextualized word representations for WSD, pre-trained on a large number of texts, namely \textit{context2vec} \cite{melamud_context2vec:_2016} and ELMo \cite{peters_deep_2018}. In these two models, WSD is treated as nearest neighbor matching as described in Section \ref{sec:incorporating_nearest}.

Table \ref{tab:results_aw} shows our WSD results in F1 measure. It is shown in the table that with the nearest neighbor matching model, BERT outperforms \textit{context2vec} and ELMo. This shows the effectiveness of BERT's pre-trained contextualized word representation. When we include surrounding sentences, one to the left and one to the right, we get improved F1 scores consistently.

We also show that linear projection to the sense output vector further improves WSD performance, by which our best results are achieved. While BERT has been shown to outperform other pre-trained contextualized word representations through the nearest neighbor matching experiments, its potential can be maximized through linear projection to the sense output vector. It is worthwhile to note that our more advanced linear projection, by means of layer weighting (\S\ref{sec:incorporating_linear_lw} and gated linear unit (\S\ref{sec:incorporating_linear_glu}) gives the best F1 scores on all test sets.

All our BERT WSD systems outperform gloss-enhanced neural WSD, which has the best overall score among all prior systems.

\subsection{English Lexical Sample Tasks}
\label{sec:results_ls}

For English lexical sample tasks, we compare our approach with the original IMS \cite{zhong_it_2010} and IMS with non-contextualized word embedding features. The embedding features incorporated into IMS include CW embeddings \cite{collobert_natural_2011}, obtained from a convolutional language model, fine-tuned (adapted) to WSD \cite{taghipour_semi-supervised_2015} (\textit{+adapted CW}), and \textit{word2vec} skip-gram \cite{mikolov_efficient_2013} with exponential decay \cite{iacobacci_embeddings_2016} (\textit{+w2v+expdecay}). We also compare our approach with the bi-LSTM, on top of which sense classification is defined \cite{kageback_word_2016}, and \textit{context2vec} \cite{melamud_context2vec:_2016}, which is a contextualized pre-trained bi-LSTM model trained on 2B words of text. Finally, we also compare with a prior multi-task and semi-supervised WSD approach learned through alternating structure optimization (ASO) \cite{ando_applying_2006}, which also utilizes unlabeled data for training.

\input{tables/results_ls.tex}

As shown in Table \ref{tab:results_ls}, our BERT-based WSD approach with linear projection model outperforms all prior approaches. \textit{context2vec}, which is pre-trained on a large amount of texts, performs worse than the prior semi-supervised ASO approach on Senseval-3, while our best result outperforms ASO by a large margin.

Neural bi-LSTM performs worse than IMS with non-contextualized word embedding features. Our neural model with pre-trained contextualized word representations outperforms the best result achieved by IMS on both Senseval-2 and Senseval-3.

\subsection{Chinese OntoNotes WSD}
\label{sec:results_chinese}

We compare our approach with IMS without and with word embedding features as the baselines. The results are shown in Table \ref{tab:results_chinese}.

\input{tables/results_chinese.tex}

Across all genres, BERT outperforms the baseline IMS with word embedding (non-contextualized word representation) features \cite{taghipour_semi-supervised_2015}. The latter also improves over the original IMS without word embedding features \cite{zhong_it_2010}. Linear projection approaches consistently outperform nearest neighbor matching by a significant margin, similar to the results on English WSD tasks.

The best overall result for the Chinese OntoNotes test set is achieved by the models with simple projection and hidden layer weighting.

\section{Discussion}
\label{sec:discussion}

Across all tasks (English all-words, English lexical sample, and Chinese OntoNotes), our experiments demonstrate the effectiveness of BERT over various prior WSD approaches. The best results are consistently obtained by linear projection models, which project the last hidden layer or the weighted sum of all hidden layers to an output sense vector.

We can view the BERT hidden layer outputs as contextual features, which serve as useful cues in determining the word senses. In fact, the attention mechanism in transformer captures the surrounding words. In prior work like IMS \cite{zhong_it_2010}, these contextual cues are captured by the manually-defined surrounding word and collocation features. The features obtained by the hidden vector output are shown to be more effective than the manually-defined features. 

We introduced two advanced linear projection techniques, namely layer weighting and gated linear unit (GLU). While \citet{peters_deep_2018} showed that the second biLSTM layer results in better WSD accuracy compared to the first layer (nearer to the individual word representation), we showed that taking into account different layers by means of the attention mechanism is useful for WSD. GLU as an activation function has been shown to be effective for better convergence and to overcome the vanishing gradient problem in the convolutional language model \cite{dauphin_language_2017}. In addition, the GLU gate vector, with values ranging from $0$ to $1$, can be seen as a filter for the features from the hidden layer vector. 

Compared with two prior contextualized word representations models, \textit{context2vec} \cite{melamud_context2vec:_2016} and ELMo \cite{peters_deep_2018}, BERT achieves higher accuracy. This shows the effectiveness of the attention mechanism used in the transformer model to represent the context.

Our BERT WSD models outperform prior neural WSD models by a large margin. These prior neural WSD models perform comparably with IMS with embeddings as classifier features, in addition to the discrete features. While neural WSD approaches \cite{kageback_word_2016,raganato_neural_2017,luo_incorporating_2018} exploit non-contextualized word embeddings which are trained on large texts, the hidden layers are trained only on a small amount of labeled data. 

\section{Conclusion}
\label{sec:conclusion}

For the WSD task, we have proposed novel strategies of incorporating BERT, a pre-trained contextualized word representation which effectively captures the context in its hidden vectors. Our experiments show that linear projection of the hidden vectors, coupled with gating to filter the values, gives better results than the prior state of the art. Compared to prior neural and feature-based WSD approaches that make use of non-contextualized word representations, using pre-trained contextualized word representation with our proposed incorporation strategy achieves significantly higher scores.

\bibliographystyle{acl_natbib}
\bibliography{wsd}

\end{document}

%% file: tables/traindata.tex
\begin{table}
\small
\centering
\begin{tabular}{|l|r|r|}
\hline
\multicolumn{1}{|c|}{\textbf{Task}} & \multicolumn{1}{c|}{\textbf{\# Instances}} & \multicolumn{1}{c|}{\textbf{\# Lexelts}} \\
\hline
English all-words & & \\
- SemCor (train) & 226,036 & 22,436 \\
- SemEval-2007 (dev) & 455   & 330 \\
- Senseval-2 (test) & 2,282 & 1,093 \\
- Senseval-3 (test) & 1,850 & 977 \\
- SemEval 2013 (test) & 1,664 & 751 \\
- SemEval 2015 (test) & 1,022 & 512 \\
\hline
English lexical sample &       &  \\
- Senseval-2 (train) & 8,611 & 177 \\
- Senseval-2 (test) & 4,328 & 146 \\
- Senseval-3 (train) & 8,022 & 57 \\
- Senseval-3 (test) & 3,944 & 57 \\
\hline
Chinese OntoNotes &       &  \\
- Train & 66,409 & 431 \\
- Dev  & 9,523 & 341 \\
- BC (test) & 1,769 & 160 \\
- BN (test) & 3,227 & 253 \\
- MZ (test) & 1,876 & 223 \\
- NW (test) & 1,483 & 143 \\
- All (test) & 8,355 & 324 \\
\hline
\end{tabular}
\caption{\label{tab:data} Statistics of the datasets used for the English all-words task, English lexical sample task, and Chinese OntoNotes WSD task in terms of the number of instances and the number of distinct lexelts. For Chinese WSD task, ``All'' refers to the concatenation of all genres BC, BN, MZ, and NW.}
\end{table}

%% file: tables/results_aw.tex
\begin{table*}[ht]
\centering
\begin{tabular}{|l|l||l|l|l|l|l|}
\hline
\multicolumn{1}{|c|}{\textbf{System}} & \multicolumn{1}{c||}{\textbf{SE07}} & \multicolumn{1}{c|}{\textbf{SE2}} & \multicolumn{1}{c|}{\textbf{SE3}} & \multicolumn{1}{c|}{\textbf{SE13}} & \multicolumn{1}{c|}{\textbf{SE15}} & \multicolumn{1}{c|}{\textbf{Avg}}\\
\hline
\multicolumn{7}{|c|}{\textit{Reported in previous papers}}\\
\hline
MFS baseline & 54.5 & 65.6 & 66.0 & 63.8 & 67.1 & 65.6\\
\hline
IMS \cite{zhong_it_2010} & 61.3 & 70.9 & 69.3 & 65.3 & 69.5 & 68.8\\
IMS+emb \cite{iacobacci_embeddings_2016} & 60.9 & 71.0 & 69.3 & 67.3 & 71.3 & 69.7\\
SupWSD \cite{papandrea_supwsd:_2017} & 60.2 & 71.3 & 68.8 & 65.8 & 70.0 & 69.0\\
SupWSD+emb \cite{papandrea_supwsd:_2017} & 63.1 & 72.7 & 70.6 & 66.8 & 71.8 & 70.5\\
\hline
BiLSTMatt+LEX \cite{raganato_neural_2017} & 63.7 & 72.0 & 69.4 & 66.4 & 72.4 & 70.1\\
GASext Concat \cite{luo_incorporating_2018} & -- & 72.2 & 70.5 & 67.2 & 72.6 & 70.6\\
\hline
context2vec \cite{melamud_context2vec:_2016} & 61.3 & 71.8 & 69.1 & 65.6 & 71.9 & 69.6\\
ELMo \cite{peters_deep_2018} & 62.2 & 71.6 & 69.6 & 66.2 & 71.3 & 69.7\\
\hline
\multicolumn{7}{|c|}{\textit{BERT nearest neighbor (ours)}}\\
\hline
1nn (1sent) & 64.0 & 73.0 & 69.7 & 67.8 & 73.3 & 71.0\\
1nn (1sent+1sur) & 63.3 & 73.8 & 71.6 & 69.2 & 74.4 & 72.3\\
\hline
\multicolumn{7}{|c|}{\textit{BERT linear projection (ours)}}\\
\hline
Simple (1sent) & 67.0 & 75.0 & 71.6 & 69.7 & 74.4 & 72.7\\
Simple (1sent+1sur) & \textbf{69.3}$^{*}$ & 75.9$^{*}$ & 73.4 & 70.4$^{*}$ & 75.1 & 73.7$^{*}$\\\hline
LW (1sent) & 66.7 & 75.0 & 71.6 & 69.9 & 74.2 & 72.7 \\
LW (1sent+1sur) & 69.0$^{*}$ & \textbf{76.4}$^{*}$ & \textbf{74.0}$^{*}$ & 70.1$^{*}$ & 75.0 & 73.9$^{*}$\\\hline
GLU (1sent) & 64.9 & 74.1 & 71.6 & 69.8 & 74.3 & 72.5 \\
GLU (1sent+1sur) & 68.1$^{*}$ & 75.5$^{*}$ & 73.6$^{*}$ & \textbf{71.1}$^{*}$ & \textbf{76.2}$^{*}$ & \textbf{74.1}$^{*}$\\\hline
GLU+LW (1sent) & 65.7 & 74.0 & 70.9 & 68.8 & 73.6 & 71.8 \\ 
GLU+LW (1sent+1sur) & 68.5$^{*}$ & 75.5$^{*}$ & 73.4$^{*}$ & 71.0$^{*}$ & \textbf{76.2}$^{*}$ & 74.0$^{*}$\\
\hline
\end{tabular}
\caption{\label{tab:results_aw} English all-words task results in F1 measure (\%), averaged over three runs. SemEval 2007 Task 17 (SE07) test set is used as the development set. We show the results of nearest neighbor matching (\textbf{1nn}) and linear projection, by \textbf{simple} last layer linear projection, layer weighting (\textbf{LW}), and gated linear units (\textbf{GLU}). Apart from BERT representation of one sentence (\textit{1sent}), we also show BERT representation of one sentence plus one surrounding sentence to the left and one to the right (\textit{1sent+1sur}). The best result in each dataset is shown in bold. Statistical significance tests by bootstrap resampling ($*$: $p<0.05$) compare 1nn (1sent+1sur) with each of Simple (1sent+1sur), LW (1sent+1sur), GLU (1sent+1sur), and GLU+LW (1sent+1sur).}
\end{table*}

%% file: tables/results_ls.tex
\begin{table}[hbt]
\small
\centering
\begin{tabular}{|l|l|l|}
\hline
\multicolumn{1}{|c|}{\textbf{System}} & \multicolumn{1}{c|}{\textbf{SE2}} & \multicolumn{1}{c|}{\textbf{SE3}} \\
\hline
\multicolumn{3}{|c|}{\textit{Reported}} \\
\hline
IMS & 65.2 & 72.3 \\
IMS+adapted CW & 66.2 & 73.4 \\
IMS+w2v+expdecay & 69.9 & 75.2 \\
\hline
BiLSTM  & 66.9 & 73.4 \\
\hline
context2vec & -- & 72.8 \\
\hline
ASO+multitask+semisup & -- & 74.1 \\
\hline
\multicolumn{3}{|c|}{\textit{BERT nearest neighbor (ours)}} \\
\hline
1nn (1sent) & 67.7 & 72.7 \\
1nn (1sent+1sur) & 71.5 & 75.7 \\
\hline
\multicolumn{3}{|c|}{\textit{BERT linear projection (ours)}} \\
\hline
Simple (1sent) & 73.3 & 78.6 \\
Simple (1sent+1sur) & \textbf{76.9}$^{*}$ & \textbf{80.0}$^{*}$ \\
LW (1sent+1sur) & 76.7$^{*}$ & \textbf{80.0}$^{*}$ \\
GLU (1sent+1sur) & 75.1$^{*}$ & 79.4$^{*}$ \\
GLU+LW (1sent+1sur) & 74.2$^{*}$ & 79.8$^{*}$ \\
\hline
\end{tabular}
\caption{\label{tab:results_ls} English lexical sample task results in accuracy (\%), averaged over three runs. Best accuracy in each dataset is shown in bold. 
Statistical significance tests by bootstrap resampling ($*$: $p<0.05$) compare 1nn (1sent+1sur) with each of Simple (1sent+1sur), LW (1sent+1sur), GLU (1sent+1sur), and GLU+LW (1sent+1sur).}
\end{table}

%% file: tables/results_chinese.tex
\begin{table}[htbp]
\small
\centering
\begin{tabular}{|l|l|l|l|l|l|}
\hline
\multicolumn{1}{|c|}{\textbf{System}} & \multicolumn{1}{c|}{\textbf{BC}} & \multicolumn{1}{c|}{\textbf{BN}} & \multicolumn{1}{c|}{\textbf{MZ}} & \multicolumn{1}{c|}{\textbf{NW}} & \multicolumn{1}{c|}{\textbf{All}} \\
\hline
\multicolumn{6}{|c|}{\textit{Baseline}} \\
\hline
IMS   & 79.8  & 85.7  & 83.0  & 91.0  & 84.8  \\
IMS+w2v & 80.2  & 86.4  & 82.3  & 92.0  & 85.2  \\
\hline
\multicolumn{6}{|c|}{\textit{BERT}} \\
\hline
1nn & 81.7  & 88.5  & 85.1  & 93.3  & 87.1  \\
\hline
Simple & 84.6$^{*}$  & \textbf{90.4}$^{*}$  & \textbf{88.9}$^{*}$ & 93.9 & \textbf{89.5}$^{*}$  \\
LW & \textbf{84.7}$^{*}$ & 90.3$^{*}$ & 88.8$^{*}$ & \textbf{94.0} & \textbf{89.5}$^{*}$  \\
GLU & 84.6$^{*}$ & 90.0$^{*}$ & 88.3$^{*}$ & 93.3 & 89.0$^{*}$ \\
GLU+LW & 84.6$^{*}$ & 90.2$^{*}$ & 88.2$^{*}$ & 93.4 & 89.2$^{*}$ \\
\hline
\end{tabular}
\caption{\label{tab:results_chinese} Chinese OntoNotes WSD results in accuracy (\%), averaged over three runs, for each genre. All BERT results in this table are obtained from the representation of one sentence plus one surrounding sentence to the left and to the right (\textit{1sent+1sur}). We show results of various BERT incorporation strategy, namely nearest neighbor matching (\textbf{1nn}), \textbf{simple} projection, projection with layer weighting (\textbf{LW}) and gated linear unit (\textbf{GLU}). Best accuracy in each genre is shown in bold.
Statistical significance tests by bootstrap resampling ($*$: $p<0.05$) compare 1nn with each of Simple, LW, GLU, and GLU+LW.}
\end{table}

%% file: wsd.bbl
\begin{thebibliography}{30}
\expandafter\ifx\csname natexlab\endcsname\relax\def\natexlab#1{#1}\fi

\bibitem[{Ando(2006)}]{ando_applying_2006}
Rie~Kubota Ando. 2006.
\newblock Applying alternating structure optimization to word sense
  disambiguation.
\newblock In \emph{Proceedings of the Tenth {Conference} on {Computational}
  {Natural} {Language} {Learning}}, pages 77--84.

\bibitem[{Ba et~al.(2016)Ba, Kiros, and Hinton}]{DBLP:journals/corr/BaKH16}
Lei~Jimmy Ba, Jamie~Ryan Kiros, and Geoffrey~E. Hinton. 2016.
\newblock Layer normalization.
\newblock \emph{CoRR}, abs/1607.06450.

\bibitem[{Bahdanau et~al.(2015)Bahdanau, Cho, and
  Bengio}]{bahdanau_neural_2015}
Dzmitry Bahdanau, Kyunghyun Cho, and Yoshua Bengio. 2015.
\newblock Neural machine translation by jointly learning to align and
  translate.
\newblock In \emph{Proceedings of the 3rd {International} {Conference} on
  {Learning} {Representations}}.

\bibitem[{Chan et~al.(2007{\natexlab{a}})Chan, Ng, and Chiang}]{chan_word_2007}
Yee~Seng Chan, Hwee~Tou Ng, and David Chiang. 2007{\natexlab{a}}.
\newblock Word sense disambiguation improves statistical machine translation.
\newblock In \emph{Proceedings of the 45th {Annual} {Meeting} of the
  {Association} for {Computational} {Linguistics}}, pages 33--40.

\bibitem[{Chan et~al.(2007{\natexlab{b}})Chan, Ng, and
  Zhong}]{chan_nus-pt:_2007}
Yee~Seng Chan, Hwee~Tou Ng, and Zhi Zhong. 2007{\natexlab{b}}.
\newblock {NUS}-{PT}: {Exploiting} parallel texts for word sense disambiguation
  in the {English} all-words tasks.
\newblock In \emph{Proceedings of the Fourth {International} {Workshop} on
  {Semantic} {Evaluations}}, pages 253--256.

\bibitem[{Collobert et~al.(2011)Collobert, Weston, Bottou, Karlen, Kavukcuoglu,
  and Kuksa}]{collobert_natural_2011}
Ronan Collobert, Jason Weston, L{\'e}on Bottou, Michael Karlen, Koray
  Kavukcuoglu, and Pavel Kuksa. 2011.
\newblock Natural language processing (almost) from scratch.
\newblock \emph{Journal of Machine Learning Research}, 12(Aug):2493--2537.

\bibitem[{Dauphin et~al.(2017)Dauphin, Fan, Auli, and
  Grangier}]{dauphin_language_2017}
Yann~N. Dauphin, Angela Fan, Michael Auli, and David Grangier. 2017.
\newblock Language modeling with gated convolutional networks.
\newblock In \emph{Proceedings of the Thirty-Fourth {International}
  {Conference} on {Machine} {Learning}}, pages 933--941.

\bibitem[{Devlin et~al.(2019)Devlin, Chang, Lee, and
  Toutanova}]{devlin_bert:_2019}
Jacob Devlin, Ming-Wei Chang, Kenton Lee, and Kristina Toutanova. 2019.
\newblock {BERT}: {Pre}-training of deep bidirectional transformers for
  language understanding.
\newblock In \emph{Proceedings of the 2019 {Conference} of the {North}
  {American} {Chapter} of the {Association} for {Computational} {Linguistics}:
  {Human} {Language} {Technologies}}, pages 4171--4186.

\bibitem[{Hochreiter and Schmidhuber(1997)}]{hochreiter_long_1997}
Sepp Hochreiter and J{\"u}rgen Schmidhuber. 1997.
\newblock Long short-term memory.
\newblock \emph{Neural Computation}, 9(8):1735--1780.

\bibitem[{Iacobacci et~al.(2016)Iacobacci, Pilehvar, and
  Navigli}]{iacobacci_embeddings_2016}
Ignacio Iacobacci, Mohammad~Taher Pilehvar, and Roberto Navigli. 2016.
\newblock Embeddings for word sense disambiguation: {An} evaluation study.
\newblock In \emph{Proceedings of the 54th {Annual} {Meeting} of the
  {Association} for {Computational} {Linguistics}}, pages 897--907.

\bibitem[{K{\r a}geb{\"a}ck and Salomonsson(2016)}]{kageback_word_2016}
Mikael K{\r a}geb{\"a}ck and Hans Salomonsson. 2016.
\newblock Word sense disambiguation using a bidirectional {LSTM}.
\newblock In \emph{Proceedings of the 5th {Workshop} on {Cognitive} {Aspects}
  of the {Lexicon}}, pages 51--56.

\bibitem[{Kingma and Ba(2015)}]{kingma_adam:_2015}
Diederik~P. Kingma and Jimmy~Lei Ba. 2015.
\newblock Adam: {A} method for stochastic optimization.
\newblock In \emph{Proceedings of the 3rd {International} {Conference} on
  {Learning} {Representations}}.

\bibitem[{Lee et~al.(2004)Lee, Ng, and Chia}]{lee_supervised_2004}
Yoong~Keok Lee, Hwee~Tou Ng, and Tee~Kiah Chia. 2004.
\newblock Supervised word sense disambiguation with support vector machines and
  multiple knowledge sources.
\newblock In \emph{Proceedings of {SENSEVAL}-3, the {Third} {International}
  {Workshop} on the {Evaluation} of {Systems} for the {Semantic} {Analysis} of
  {Text}}, pages 137--140.

\bibitem[{Luo et~al.(2018)Luo, Liu, Xia, Chang, and
  Sui}]{luo_incorporating_2018}
Fuli Luo, Tianyu Liu, Qiaolin Xia, Baobao Chang, and Zhifang Sui. 2018.
\newblock Incorporating glosses into neural word sense disambiguation.
\newblock In \emph{Proceedings of the 56th {Annual} {Meeting} of the
  {Association} for {Computational} {Linguistics}}, pages 2473--2482.

\bibitem[{Ma and Hovy(2016)}]{ma_end--end_2016}
Xuezhe Ma and Eduard Hovy. 2016.
\newblock End-to-end sequence labeling via bi-directional {LSTM}-{CNNs}-{CRF}.
\newblock In \emph{Proceedings of the 54th {Annual} {Meeting} of the
  {Association} for {Computational} {Linguistics},}, pages 1064--1074.

\bibitem[{McCann et~al.(2017)McCann, Bradbury, Xiong, and
  Socher}]{mccann_learned_2017}
Bryan McCann, James Bradbury, Caiming Xiong, and Richard Socher. 2017.
\newblock Learned in translation: {Contextualized} word vectors.
\newblock In \emph{Proceedings of the Thirty-First {Conference} on {Neural}
  {Information} {Processing} {Systems}}, pages 6297--6308.

\bibitem[{Melamud et~al.(2016)Melamud, Goldberger, and
  Dagan}]{melamud_context2vec:_2016}
Oren Melamud, Jacob Goldberger, and Ido Dagan. 2016.
\newblock context2vec: {Learning} generic context embedding with bidirectional
  {LSTM}.
\newblock In \emph{Proceedings of the 20th {SIGNLL} {Conference} on
  {Computational} {Natural} {Language} {Learning}}, pages 51--61.

\bibitem[{Mikolov et~al.(2013)Mikolov, Chen, Corrado, and
  Dean}]{mikolov_efficient_2013}
Tomas Mikolov, Kai Chen, Greg Corrado, and Jeffrey Dean. 2013.
\newblock Efficient estimation of word representations in vector space.
\newblock In \emph{Proceedings of {Workshop} at the {International}
  {Conference} on {Learning} {Representations}}.

\bibitem[{Miller et~al.(1994)Miller, Chodorow, Landes, Leacock, and
  Thomas}]{miller_using_1994}
George~A. Miller, Martin Chodorow, Shari Landes, Claudia Leacock, and Robert~G.
  Thomas. 1994.
\newblock Using a semantic concordance for sense identification.
\newblock In \emph{Human Language Technology: {Proceedings} of a {Workshop}
  held at {Plainsboro}, {New} {Jersey}}, pages 240--243.

\bibitem[{Papandrea et~al.(2017)Papandrea, Raganato, and
  Delli~Bovi}]{papandrea_supwsd:_2017}
Simone Papandrea, Alessandro Raganato, and Claudio Delli~Bovi. 2017.
\newblock {SupWSD}: {A} flexible toolkit for supervised word sense
  disambiguation.
\newblock In \emph{Proceedings of the 2017 {Conference} on {Empirical}
  {Methods} in {Natural} {Language} {Processing}: {System} {Demonstrations}},
  pages 103--108.

\bibitem[{Peters et~al.(2018)Peters, Neumann, Iyyer, Gardner, Clark, Lee, and
  Zettlemoyer}]{peters_deep_2018}
Matthew~E. Peters, Mark Neumann, Mohit Iyyer, Matt Gardner, Christopher Clark,
  Kenton Lee, and Luke Zettlemoyer. 2018.
\newblock Deep contextualized word representations.
\newblock In \emph{Proceedings of the 2018 {Conference} of the {North}
  {American} {Chapter} of the {Association} for {Computational} {Linguistics}:
  {Human} {Language} {Technologies}}, pages 2227--2237.

\bibitem[{Pradhan et~al.(2013)Pradhan, Moschitti, Xue, Ng, Bj{\"o}rkelund,
  Uryupina, Zhang, and Zhong}]{pradhan_towards_2013}
Sameer Pradhan, Alessandro Moschitti, Nianwen Xue, Hwee~Tou Ng, Anders
  Bj{\"o}rkelund, Olga Uryupina, Yuchen Zhang, and Zhi Zhong. 2013.
\newblock Towards robust linguistic analysis using {OntoNotes}.
\newblock In \emph{Proceedings of the {Seventeenth} {Conference} on
  {Computational} {Natural} {Language} {Learning}}, pages 143--152.

\bibitem[{Raganato et~al.(2017{\natexlab{a}})Raganato, Camacho-Collados, and
  Navigli}]{raganato_word_2017}
Alessandro Raganato, Jose Camacho-Collados, and Roberto Navigli.
  2017{\natexlab{a}}.
\newblock Word sense disambiguation: {A} unified evaluation framework and
  empirical comparison.
\newblock In \emph{Proceedings of the 15th {Conference} of the {European}
  {Chapter} of the {Association} for {Computational} {Linguistics}}, pages
  99--110.

\bibitem[{Raganato et~al.(2017{\natexlab{b}})Raganato, Delli~Bovi, and
  Navigli}]{raganato_neural_2017}
Alessandro Raganato, Claudio Delli~Bovi, and Roberto Navigli.
  2017{\natexlab{b}}.
\newblock Neural sequence learning models for word sense disambiguation.
\newblock In \emph{Proceedings of the 2017 {Conference} on {Empirical}
  {Methods} in {Natural} {Language} {Processing}}, pages 1156--1167.

\bibitem[{Sutskever et~al.(2014)Sutskever, Vinyals, and
  Le}]{sutskever_sequence_2014}
Ilya Sutskever, Oriol Vinyals, and Quoc~V. Le. 2014.
\newblock Sequence to sequence learning with neural networks.
\newblock In \emph{Proceedings of the Twenty-Eighth {Conference} on {Neural}
  {Information} {Processing} {Systems}}, pages 3104--3112.

\bibitem[{Taghipour and Ng(2015)}]{taghipour_semi-supervised_2015}
Kaveh Taghipour and Hwee~Tou Ng. 2015.
\newblock Semi-supervised word sense disambiguation using word embeddings in
  general and specific domains.
\newblock In \emph{Human {Language} {Technologies}: {The} 2015 {Annual}
  {Conference} of the {North} {American} {Chapter} of the {ACL}}, pages
  314--323.

\bibitem[{Turian et~al.(2010)Turian, Ratinov, and Bengio}]{turian_word_2010}
Joseph Turian, Lev Ratinov, and Yoshua Bengio. 2010.
\newblock Word representations: {A} simple and general method for
  semi-supervised learning.
\newblock In \emph{Proceedings of the 48th {Annual} {Meeting} of the
  {Association} for {Computational} {Linguistics}}, pages 384--394.

\bibitem[{Vaswani et~al.(2017)Vaswani, Shazeer, Parmar, Uszkoreit, Jones,
  Gomez, Kaiser, and Polosukhin}]{vaswani_attention_2017}
Ashish Vaswani, Noam Shazeer, Niki Parmar, Jakob Uszkoreit, Llion Jones,
  Aidan~N. Gomez, {\L }ukasz Kaiser, and Illia Polosukhin. 2017.
\newblock Attention is all you need.
\newblock In \emph{Proceedings of the Thirty-First {Conference} on {Neural}
  {Information} {Processing} {Systems}}, pages 5998--6008.

\bibitem[{Zhong and Ng(2010)}]{zhong_it_2010}
Zhi Zhong and Hwee~Tou Ng. 2010.
\newblock It {Makes} {Sense}: a wide-coverage word sense disambiguation system
  for free text.
\newblock In \emph{Proceedings of the {ACL} 2010 {System} {Demonstrations}},
  pages 78--83.

\bibitem[{Zhong and Ng(2012)}]{zhong_word_2012}
Zhi Zhong and Hwee~Tou Ng. 2012.
\newblock Word sense disambiguation improves information retrieval.
\newblock In \emph{Proceedings of the 50th {Annual} {Meeting} of the
  {Association} for {Computational} {Linguistics}}, pages 273--282.

\end{thebibliography}
